\documentclass{article}

\usepackage[final]{corl_2020} % Uncomment for the camera-ready ``final'' version.

\usepackage{titlesec}
\titlespacing*{\section}
{0pt}{.4\baselineskip}{0.5\baselineskip}
\titlespacing*{\subsection}
  {0pt}{0.25\baselineskip}{0.2\baselineskip}

\title{Learning a Contact-Adaptive Controller for \\Robust, Efficient Legged Locomotion}

\author{%
Xingye Da$^{*}$, %
Zhaoming Xie$^{*\dagger}$, % 
David Hoeller$^{*}$, %
Byron Boots$^{*\ddagger}$, %, %
\\
\textbf{Animashree Anandkumar$^{*\mathsection}$, %
Yuke Zhu$^{*\sharp}$, %
Buck Babich$^{*}$, %
Animesh Garg$^{\star}$}%
\thanks{NVIDIA, %
$^{\dagger}$Univ.\ of British Columbia, %
$^{\ddagger}$Univ.\ of Washington, %
$^{\mathsection}$Caltech, % 
$^{\sharp}$UT Austin, %
$^{\star}$Univ.\ of Toronto, Vector Institute. 
Work done at NVIDIA. Correspondence: \texttt{xda@nvidia.com, zxie47@cs.ubc.ca}} 
}

\begin{document}
\maketitle

%===============================================================================

\begin{abstract}
    We present a hierarchical framework that combines model-based control and reinforcement learning (RL) to synthesize robust controllers for a quadruped (the Unitree Laikago). The system consists of a high-level controller that learns to choose from a set of primitives in response to changes in the environment and a low-level controller that utilizes an established control method to robustly execute the primitives. Our framework learns a controller that can adapt to challenging environmental changes on the fly, including novel scenarios not seen during training. The learned controller is up to 85~percent more energy efficient and is more robust compared to baseline methods. We also deploy the controller on a physical robot without any randomization or adaptation scheme.
\end{abstract}


\section{Introduction}

% \textbf{DRAFT: The intro will be heavily edited later today (Monday).}

\begin{figure}
    \centering
    \includegraphics[width=0.8\columnwidth]{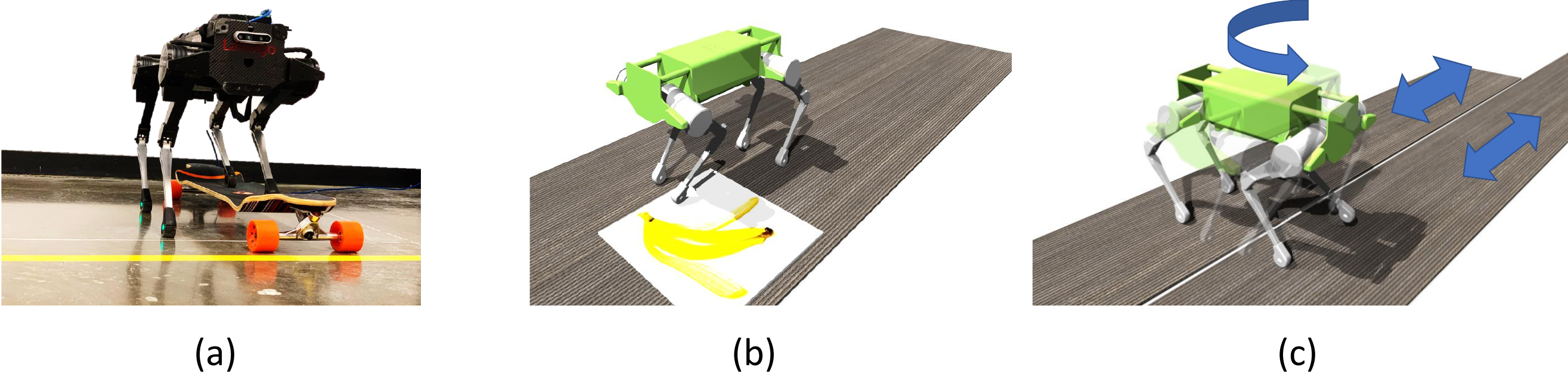}
    \caption{(a) Riding a skateboard requires a contact sequence that only moves the feet on the ground while keeping the feet on the board still. (b) "banana peel" test: we put a frictionless mat under a foot to test robustness. (c) We train and test the robot on a split-belt treadmill where the speeds of the two belts are changed separately with the robot facing different directions.}
    \label{fig:first photo}
    \vspace{-15pt}
\end{figure}

Quadruped locomotion is often characterized in terms of \emph{gaits} (walking, trotting, galloping, bounding, etc.) that have been well-studied in animals \cite{1981-nature-horseGait} and reproduced on robots~\cite{2018-iros-cheetahMPC, RoboImitationPeng20}. A gait is a periodic contact sequence that defines a specific contact timing for each foot. Controllers designed for these gaits have demonstrated robust behaviors on flat ground and rough terrain locomotion. However, it is rarer to find controllers that can change gaits or contact sequences to adapt to environmental changes. An adaptive gait can reduce energy usage by removing unnecessary movement, as suggested in horse studies~\cite{1981-nature-horseGait}. It is also required for completing more challenging scenarios such as riding a skateboard or recovery from leg slipping, as shown in Figure~\ref{fig:first photo} (a, b).
% a more sophistic contact sequence design is required.

In most model-based and learning-based control designs, the contact sequence is fixed or predefined~\cite{2018-iros-cheetahMPC, 2017-humanoids-hyqMPC, 2019-RAL-towr, 2020-icra-crocoddyl, RoboImitationPeng20, 2018-rss-sim2realquadruped, 2019-science-sim2realAnyaml}. Dynamic adaptation of the contact sequence is challenging because of the hybrid nature of legged locomotion dynamics as well as the inherent instability of such systems. While it is possible to generate adaptive contact schemes via trajectory optimization~\cite{2014-ijrr-direct_traj, 2020-springer-zacContactImplicit, 2012-ACM-Igor_contact_invariant_opti}, such approaches are generally too compute-intensive for real-time use. 

Here we present a hierarchical control framework for quadrupedal locomotion that learns to adaptively change contact sequences in real-time. A high-level controller is trained with reinforcement learning (RL) to specify the contact configuration of the feet, which is then taken as input by a low-level controller to generate ground reaction forces via quadratic programming (QP). At inference time, the high-level controller needs only evaluate a small multi-layer neural network, avoiding the use of an expensive model predictive control (MPC) strategy that might otherwise be required to optimize for long-term performance. The low-level controller provides high-bandwidth feedback to track base and foot positions and also helps ensure that learning is sample-efficient. The framework produces a controller that is up to 85~percent more energy efficient and also more robust than baseline approaches.
% It also makes the controller easier to generalize to new scenarios without extra data collection.

We train our controller with a simulated Unitree Laikago~\cite{laikago} on a split-belt treadmill, as shown in Figure~\ref{fig:first photo} (c). The two belts can adjust speed independently, and we change the robot orientation to increase variation. In addition to comparing energy use and robustness to the baselines, we also demonstrate zero-shot transferability by testing the controller in novel situations such as one where a foot encounters a slippery surface (e.g., with zero friction), which we call the ``banana peel" test. Furthermore, we show that the controller learns to generate novel contact sequences that have not been previously shown in either the model-based or learning-based approaches. Finally, we deploy the controller on a physical robot to demonstrate sim-to-real transfer,\footnote{Due to the COVID-19 pandemic, access to the physical robot has been limited. We thus focus our quantitative analysis on simulation while demonstrating only qualitative results on the physical robot. \\ Video \url{https://youtu.be/JJOmFZKpYTo} \\ Website \url{https://sites.google.com/view/learn-contact-controller/home}} 
% both in the paper and supplementary video.} 
which succeeds without any randomization or adaptation scheme due to the robustness of the low-level controller.

\noindent \textbf{Summary of Contributions}:
% \begin{itemize}[leftmargin=*]
\begin{enumerate}[
    topsep=0pt,
    noitemsep,
    % partopsep=1ex,
    % parsep=1ex,
    leftmargin=*,
    % itemindent=3ex
    ]
    \item We introduce a hierarchical control structure that combines model-based control design and model-free reinforcement learning for legged locomotion.
    \item We demonstrate that our framework allows sample-efficient learning, zero-shot adaptation to novel scenarios, and direct sim-to-real transfer without randomization or adaptation schemes.
    \item Our framework learns adaptive contact sequences that are not present in either model-based or learning-based methods in real-time control. This is evidenced in the natural-looking behaviors that minimize unnecessary movement and energy usage in our split-belt treadmill scenarios.
    % \item The framework, we can get robust controller. We define motion primitives. We train high-level policy that can select the multi task
    % \item More robust than heuristic design, and end-to-end. Measurement, continuous measurement, 
    % \item More sample efficient.
    % \item Show a set of experiments. Each of the experiment shows a particular behavior.
    % \item Natural-looking behaviors that minimize the unnecessary movement and reduce the energy use.
\end{enumerate}

\section{Related Work}

\paragraph{Model-based Legged Locomotion Control} Model-based control designs \cite{2018-iros-cheetahMPC, 2017-humanoids-hyqMPC, 2019-RAL-towr, 2020-icra-crocoddyl} use trajectory optimization and model predictive control methods that optimize the performance for a finite horizon, where the input includes a predefined contact sequence. Although one can change the control sequence externally to demonstrate various gaits, it cannot adapt to changes in the environment. Contact-implicit optimization \cite{2014-ijrr-direct_traj, 2020-springer-zacContactImplicit, 2012-ACM-Igor_contact_invariant_opti} is used to solve non-convex, stiff problems that are not amenable to real-time use. The work in \cite{2019-RAL-impulseSet} introduced the Feasible Impulse Set which allows online gait adaptation in planar models, but no 3D work has been presented.

\paragraph{Learning Legged Locomotion}
Recently there has been significant work investigating the use of reinforcement learning to obtain locomotion policies for legged robots \cite{2018-rss-sim2realquadruped, RoboImitationPeng20, 2019-CORL-cassie, 2019-science-sim2realAnyaml}. However, the resulting policies are often either less robust compared to controllers obtained from model-based methods (e.g., \cite{2018-rss-sim2realquadruped, RoboImitationPeng20}) or require the design of complicated reward functions and a large number of training samples, e.g \cite{2019-science-sim2realAnyaml}. A learned policy is often brittle and can fail under mild environmental changes. Many approaches like meta learning \cite{2020-icra-meta_adapt} or Bayesian Optimization in behavior space \cite{2015-nature-adapt} have been proposed to adaptively update the policies. This usually requires the policy to interact with the target environment to collect additional data. In contrast to these approaches, our method can adapt to a changing environment without any online data collection.

\paragraph{Hierarchical Control} Hierarchical framework can greatly improve learning efficiency, as shown in many prior works, e.g., \cite{2001-precup_thesis-temporal_option}. In robotics tasks, it is beneficial to decompose a controller into modules and obtain controllers in a hierarchical manner. A low-level controller can be model-based \cite{2020-icra-hierachicalManipulation, 2016-Springer-switch_primitive,  2016-TOG-terrainRL} or learned \cite{2019-corl-hierachicalsim2real, 2017-TOG-deepLoco} such that it can achieve subgoals specified by a learned high-level controller. Specifically for locomotion tasks, it is natural to decompose a controller into direction-following and navigation modules \cite{2017-TOG-deepLoco, 2019-corl-hierachicalsim2real, 2018-nips-hierachicalRL, 2019-arxiv-hierarchicalQuadruped, 2020-icra-deepgait}.
We also adopt a hierarchical structure, however, our work is distinguished from previous work in that the goal of the high-level controller is to choose a low-level primitive in order to adapt to environmental changes instead of specifying subgoals. This does not preclude other high-level policies or behaviors; e.g.. one could easily add a navigation module in our framework. 

\section{Method}

\begin{figure}
    \centering
    \includegraphics[width=0.8\columnwidth]{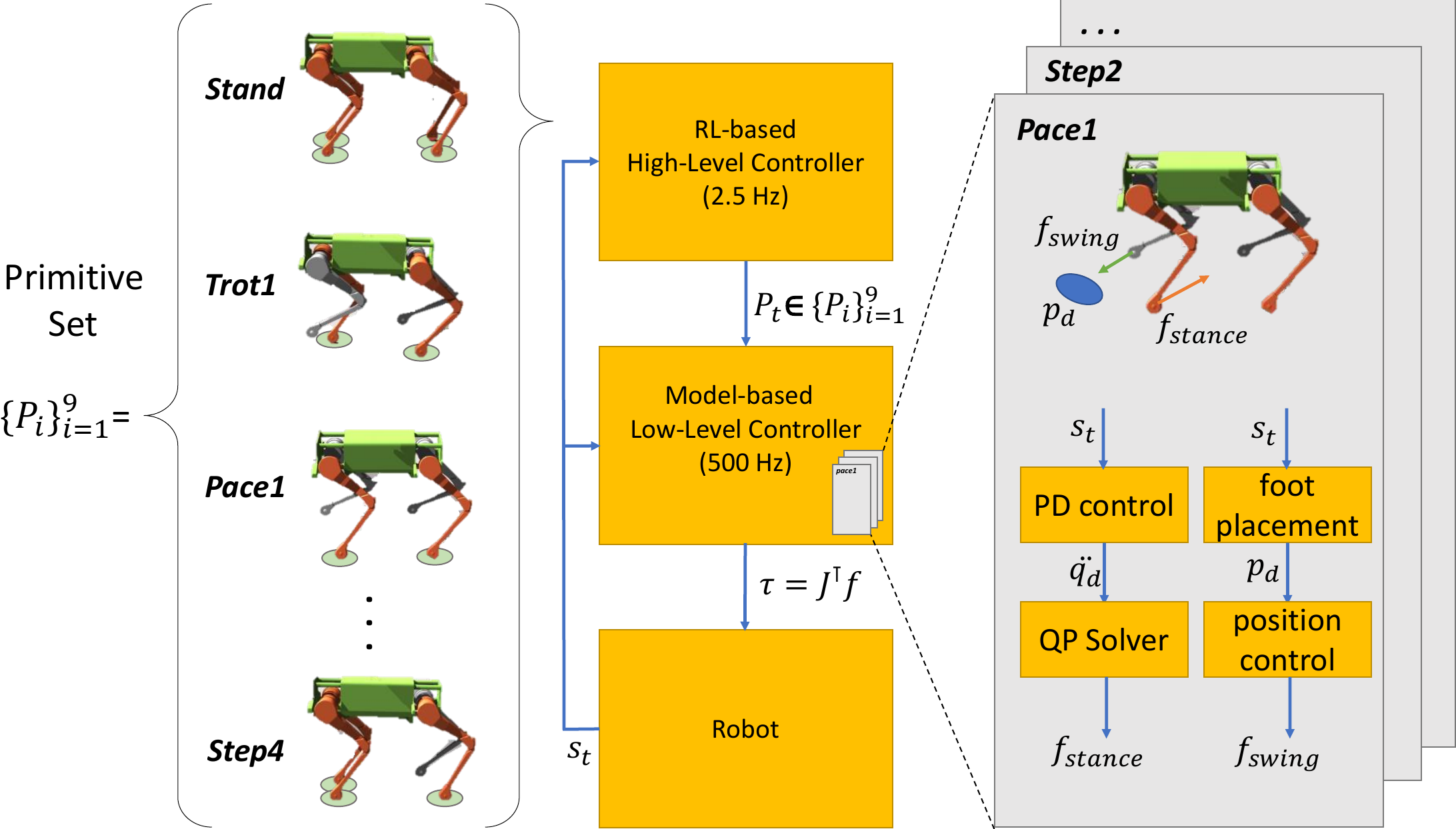}
    \caption{Overview of our system. \textbf{Left}: Primitives $P_i$ are distinguished by the contact configuration. The stance legs in each primitive are colored orange. \textbf{Center}: Hierarchical structure of the controller. The high-level controller chooses from a set of primitives based on the robot state $s_t$, and the low-level controller computes the motor torques $\tau$ based on the robot state and the primitive chosen. \textbf{Right}: The low-level controller uses stance foot forces to control the base pose and moves the swing feet to their target positions.}
    \label{fig:system}
    % \vspace{-20pt}
\end{figure}

We propose a hierarchical framework to perform locomotion that combines model-based control with reinforcement learning. The system is visualized in Figure~\ref{fig:system}.

The state of the robot $s = (q, \dot{q}, p_\text{foot})$ consists of the base pose $q \in \mathbb R^3 \times S^3$, containing the position $p_\text{body}$ and orientation $\Theta$ of the robot's body, the velocity vector $\dot{q} \in \mathbb R^6$, and the four Cartesian foot positions $p_\text{foot} = (p_1, p_2, p_3, p_4) \in \mathbb R^{12}$ relative to the base. A primitive $P =  \{0:\text{Stance}, 1:\text{Swing}\}^4 \in \mathbb{Z}^4$ is a Boolean array that specifies the stance/swing state of the four feet. The high-level controller chooses the appropriate primitive and the low-level controller uses the stance feet to generate ground reaction forces by solving a QP for base pose control and moves the swing feet based on a foot-placement algorithm. The low-level controller runs at $\SI{500}{\hertz}$ for high-speed feedback and the high-level controller runs at $\SI{2.5}{\hertz}$ to match the primitive execution time. 

The high-level controller can be designed manually. If we set all foot states to \textit{Stance}, the result is a standing gait. If we instead synchronize diagonally opposed feet and switch the state at every time step, the result is the trotting gait. As we demonstrate, however, a more adaptive high-level controller is needed to reduce energy consumption, reject disturbance, or react to friction changes.

In this section, we first define the primitives that are used in the controller. We then give details on the low-level controller that implements these primitives and on the high-level controller that learns to select from these primitives to complete multiple tasks. 

\subsection{Primitives}
A primitive $P$ represents a contact configuration for the four feet. Each foot is in either a \textit{Stance} or \textit{Swing} state, and there are thus $2^4=16$ possible primitives in total. We only use $9$ primitives that are commonly used in quadruped locomotion, as summarized in the following table: 
% \vspace{-5pt}
\begin{center}
\begin{tabular}{c|ccccc}
\hline
 Primitive & \textbf{\textit{Stand}} & \textbf{\textit{Trot1}} & \textbf{\textit{Trot2}} & \textbf{\textit{Pace1}} & \textbf{\textit{Pace2}} \\ \hline
 Feet State & [0 0 0 0] & [1 0 0 1] & [0 1 1 0] & [0 1 0 1] & [1 0 1 0] \\ \hline \hline
 Primitive & \textbf{\textit{Step1}} & \textbf{\textit{Step2}} & \textbf{\textit{Step3}} & \textbf{\textit{Step4}} \\ \hline
 Feet State & [1 0 0 0] & [0 1 0 0] & [0 0 1 0] & [0 0 0 1] \\ \hline
\end{tabular}
\end{center}
where $0$ indicates that the corresponding foot is a stance foot and $1$ indicates otherwise. The order of the foot states is $\{\text{Left Front, Right Front, Left Rear, Right Rear} \}$ or $\{\text{LF, RF, LR, RR}\}$ in short.

\subsection{Low-Level Controller}
\label{sec:low_level}
We implement each primitive with a low-level torque controller. We find that a simple model-based method is sufficient to complete most of our tasks and is straightforward to transfer to the real robot.

\paragraph{Base Pose Control}
The low-level controller receives a primitive $P_t$ from the high-level controller. It also receives the target base pose $q_d$ and velocity $\dot{q}_d$ from user command. The controller computes foot forces by solving a QP, so the base pose can track the target pose and respect contact constraints. 

Similar to \cite{2018-iros-cheetahMPC}, we approximates the quadruped dynamics as a linearized centroidal dynamics,

\begin{equation}
\ddot{q} = \mathbf{M} f-\tilde{g},
\end{equation}

where $\mathbf{M} \in \mathbb{R}^{6\times12}$ is the inverse inertia matrix, $f = (f_1, f_2, f_3, f_4) \in \mathbb{R}^{12}$ is the column vector of Cartesian forces for each foot, and $\tilde{g} = (g, 0_3) \in \mathbb{R}^6$ is the augmented gravity vector. The detailed derivation is given in Appendix~\ref{sec:dynamics derivation}. 

Given a target base pose $q_d$ and velocity $\dot{q}_d$, we use PD control to compute the target acceleration
\begin{equation}
\ddot{q}_d = k_p(q_{d}-q)+k_d(\dot{q}_d - \dot{q}).
\end{equation}
This is then used to construct a QP to find foot forces that minimize the acceleration error while respecting the contact configuration and friction constraints
\begin{equation}
\begin{aligned}
 \underset{f}{\text{min}}
\quad & ||\mathbf{M}f - \tilde{g} - \ddot{q}_d||_\mathbf{Q} + ||f||_\mathbf{R} \\
 \text{subject to}
\quad & f_{z, i} \geq f_{z, \text{min}} \,\qquad \text{if } P_{t, i} \text{ is } \bf{Stance} \\
\quad & f_{z, i} = 0 \qquad\qquad \text{if } P_{t, i} \text{ is } \bf{Swing} \\
& -\mu\; f_{x} \leq f_{z} \leq \mu\; f_{x} \\
& -\mu\; f_{y} \leq f_{z} \leq \mu\; f_{y},
\end{aligned}
\label{eq:QP}
\end{equation}
where $\mathbf{Q}$ and $\mathbf{R}$ are diagonal matrices that adjust weights in the cost function.

\paragraph{Swing Foot Control}

The desired foot position $p_{d, i}$ for foot $i$ is computed by a linear foot-placement heuristic
\begin{equation}
    p_{d, i} = p_{0, i} + k (\dot{p}_{\text{body}} -  \dot{p}_{d, \text{body}}),
\end{equation}
that adjusts position from the default state $p_{0, i}$.

A position controller is then used to compute the swing foot force by
\begin{equation}
    f_i = k_{p,i} (p_{d, i} -  p_i)-k_{d, i}\dot{p}_i.
\end{equation}

% \subsubsection{Torque Control}
\paragraph{Torque Control} The foot forces computed via pose control and swing foot control are converted to motor torques by $\tau = J^T f$, where $J \in \mathbb{R}^{12\times12}$ is the feet positions Jacobian matrix with respect to motor states. This is updated at \SI{500}{\hertz}.

% \begin{figure}
%     \centering
%     \includegraphics[width=0.8\columnwidth]{figure/low_level_controller.pdf}
%     \caption{The low level controller computes foot forces based on contact configuration. The stance foot forces are computed via QP to regulate body pose and the swing foot forces are computed via position control. The foot forces are transformed into joint torques via Jacobian transpose.}
%     \label{fig:low_level_control}
% \end{figure}

% \subsection{Gait Selection Policy}
\subsection{High-level Controller}
\label{sec:high_level}
Our high-level controller selects primitives based on the current robot state and is queried at \SI{2.5}{\hertz}. Here we describe how we use RL to learn the high-level controller in detail.

\paragraph{State Space and Action Space} We model the environment as a partially observable Markov decision process (POMDP). Specifically, the high-level controller takes the body pose $\bf{q}$, excluding the $x, y$ linear positions, and the relative foot positions $\bf{p_\textit{foot}}$ as input. To endow the controller with the capability to learn common gaits such as pacing and trotting that alternate between primitives, we also include the previously-used primitive as an input. The output of the controller is a $9$-dimensional one-hot vector that indicates which primitive will be selected for the low-level controller. Assuming the environment is deterministic, the input provided is enough to determine the next robot state. However, environments are often parameterized by some unobserved random variables, causing the transition dynamics to be stochastic with high variance. The goal of the high-level controller is then to choose primitives that can adapt to this high variance environment while optimizing for simple objectives such as energy efficiency and stability. 

\paragraph{Policy Representation} Since the action space is discrete, instead of learning a policy directly, we choose to learn a Q-function that takes the state and action as input, and output the sum of discounted future rewards. At test time, the action that yields the maximum Q value is selected.

\paragraph{Reward Design} We use a simple reward function of the form  
\begin{equation}
r = 1 - 0.0025\frac{1}{T}\sum \norm{\tau}^2 - \frac{1}{T}\sum\norm{\dot{p}_{d, \text{body}} -  \dot{p}_\text{body}}^2,  
\end{equation}
where the constant $1$ ensures that the reward is positive, $\tau, {p_{d, \text{body}}}$ and $\dot{p}_\text{body}$ are control torques, body linear velocity, and desired body linear velocity respectively. $T$ is the number of simulation timesteps within a primitive cycle.

\paragraph{Training} We adopt a DQN-like \cite{2015-nature-dqn} training procedure, and implement a double Q-function \cite{2016-aaai-doubleDQN} and delayed target network update \cite{2018-icml-td3}. In addition, instead of using an epsilon greedy strategy during training, the probability of applying an action is based on Q value estimates normalized by the softmax operator. More specifically, let $\{Q_1, Q_2, \dots, Q_9\}$ be the Q value estimates of the different actions at a particular state, and let the maximum Q value be $Q_{\max} = \max(Q_1, Q_2, \dots, Q_9)$, the probability of an action $i$ being sampled will be proportional to $\exp(-\nu \frac{Q_i}{Q_{\max}})$, where $\nu$ is a hyperparameter controlling how sensitive the sampling probability distribution is to the Q value. The pseudocode of the algorithm is described in Appendix~\ref{sec:DQN}.

The Q-function for the high-level policies is implemented as a two-layer feedforward neural network with ReLU activation functions, each layer has $64$ neurons. We set the temperature $\nu = 5$ and update the Q-function every $100$ samples, with $50$ stochastic gradient descents and mini-batch size of $512$. The learned policy already performs well with $10^5$ samples, and we collect a maximum of $5\times 10^5$ samples. Note that this is orders of magnitude fewer samples than used in previous work that employed RL with similar-scale quadrupeds like the ANYmal or Laikago \cite{2019-science-sim2realAnyaml, RoboImitationPeng20, 2020-icra-constraintQuadruped}.

\section{Results}
\label{sec:results}

\begin{figure}
    \centering
    \includegraphics[width=0.98\columnwidth]{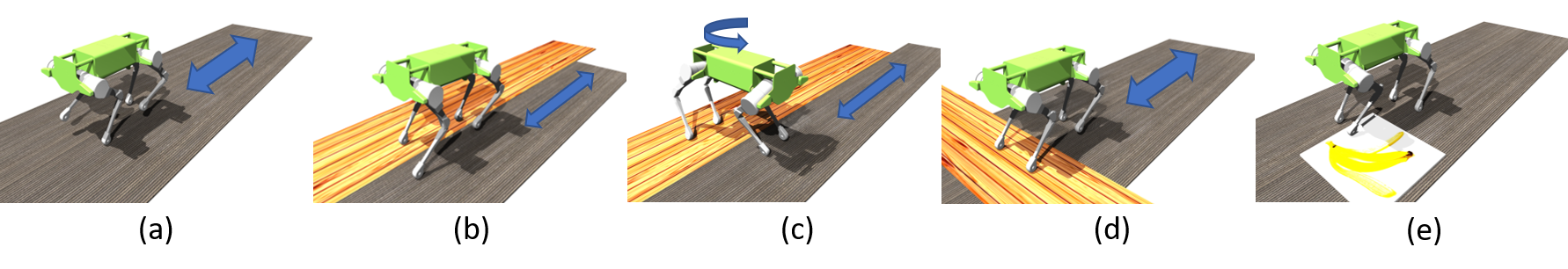}
    \caption{Training and Testing scenarios. Scenarios (a)-(c) scenarios are used during training where we vary the treadmill speeds, the number of moving belts, and the orientation of the robot. Scenarios (d)-(e) are introduced only during testing. Scenario (d) introduces a fixed plywood bridge on top of the treadmill, and scenario (e) inserts a frictionless mat under the feet of the robot to test stability.}
    \label{fig:test_scenario}
    % \vspace{-15pt}
\end{figure}

We use a GPU-accelerated simulator \cite{2020-IsaacGym} Isaac Gym to train the Laikago robot and to compare different controllers. This simulator has been used for various robotics manipulation sim-to-real tasks \cite{2020-arxiv-phyx1, 2019-icra-phyx2} and is validated to simulate rigid body physics with reasonable accuracy. In this section, we describe the experiments we use to validate our framework and show improvement over baseline methods.

\begin{figure}
    \centering
    \begin{subfigure}[b]{0.48\columnwidth}
        \centering
        \includegraphics[width=0.98\columnwidth]{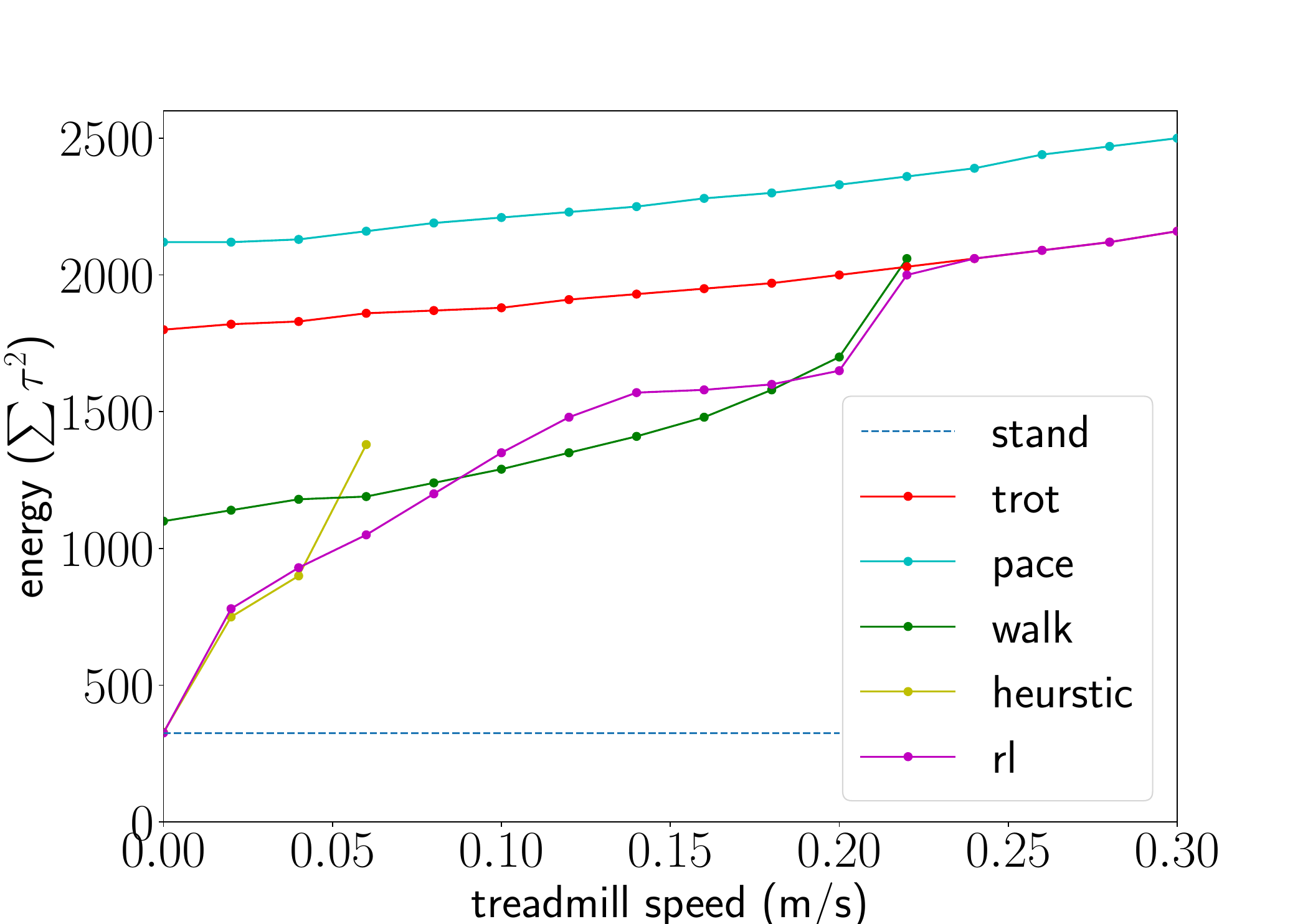}
        \caption{Energy comparison over different speeds.}
    \end{subfigure}%
    ~ 
    \begin{subfigure}[b]{0.48\columnwidth}
        \centering
        \includegraphics[width=0.98\columnwidth]{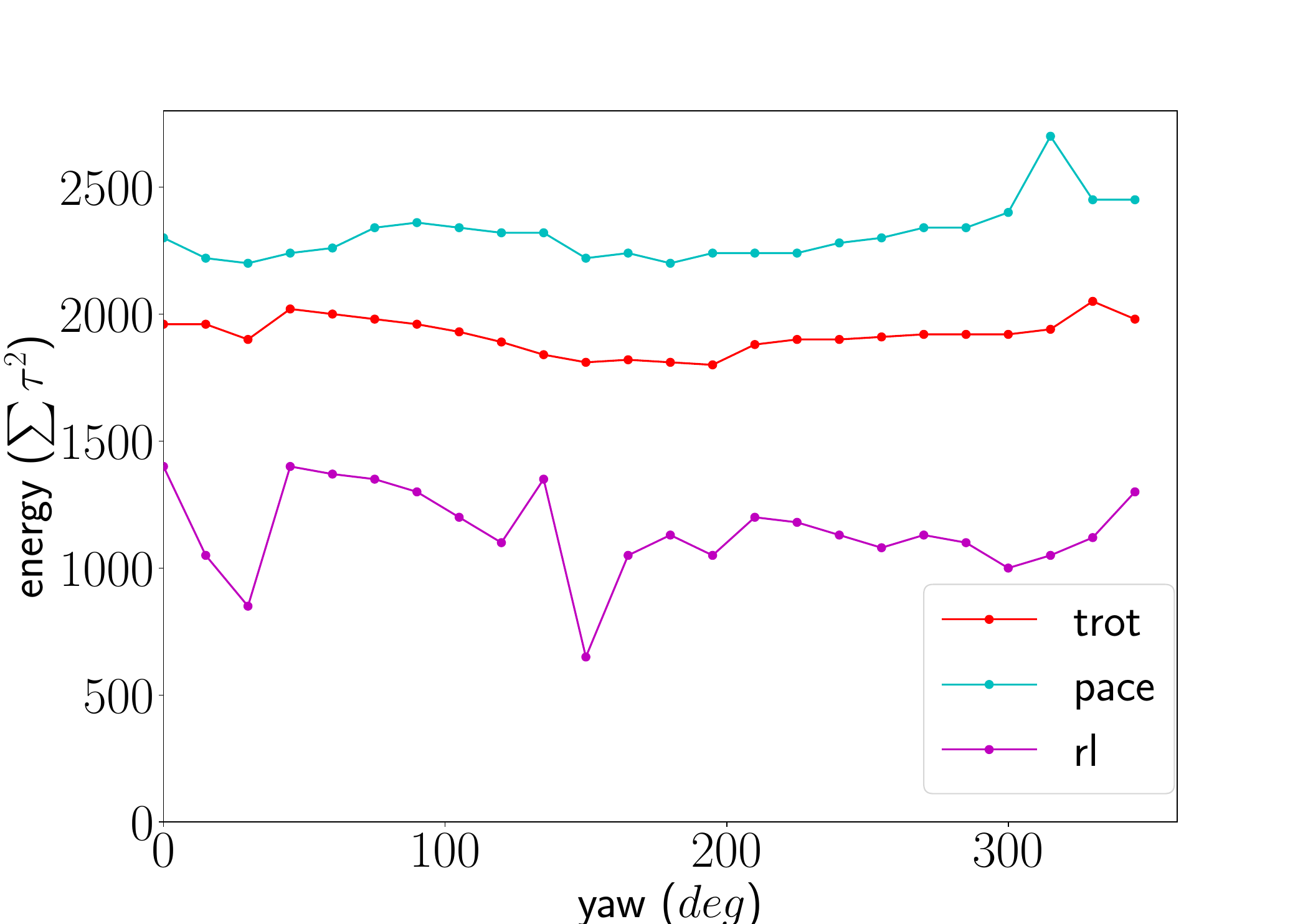}
        \caption{Energy comparison over different yaws.}
    \end{subfigure}%
    \caption{Comparison of the average energy used. (a) The standing, walking and heuristic controllers fails at high speed, while trotting and pacing controllers remain on high-energy level. The learned controller (rl) can handle all speed variation and more energy efficient. (b) The only baseline controllers that can handle split-treadmill are trotting and pacing. The learned controller is 50~percent more energy efficient on average. Energy for the learned controller drops significantly at $yaw = 150~\text{deg}$ because only one foot moves while two feet move in nearby orientations.}
    \label{fig:energy_comparison}
    % \vspace{-20pt}
\end{figure}

\subsection{Baseline Controllers}
% \textbf{Included controllers}
% \begin{itemize}
%     \item Standing
%     \item Trotting
%     \item Placing
%     \item Walking
% \end{itemize}

% Measurement: tracking the command speed. reject initial speed perturbation. Energy level for different speed.

% Note: The standing gait will have the lowest energy in zero speed, but other gaits have a better tracking. This will motive us to use adaptive transition. 

% Explanation: The motion primitives can be used to generate versatile of controller. For example, combine two trotting primitives to get trotting gait, and combine four stepping primitives to get walking gait. These gaits can achieve a few tasks, like walking forward, turning.

As a baseline, we have created five manually-designed high-level controllers. These are comparable to those used in typical model-based control approaches to quadrupedal locomotion \cite{2017-humanoids-hyqMPC,2018-iros-cheetahMPC}. Here we briefly describe how they work.

\paragraph{Standing} Only the \textbf{\textit{Stand}} primitive is used in this controller. It is the most energy efficient controller in the absence of perturbations but also the least robust one if perturbations are present.

\paragraph{Trotting} The trotting controller alternates between \textbf{\textit{Trot1}} and \textbf{\textit{Trot2}} and displays a trotting gait. It is commonly used for quadruped locomotion due to its stability.

\paragraph{Pacing} The pacing controller alternates between \textbf{\textit{Pace1}} and \textbf{\textit{Pace2}}. It is another commonly used gait in quadrupedal locomotion but usually less stable and less energy efficient than trotting.

\paragraph{Walking} The walking controller lifts one foot at a time by switching between four stepping primitives in the order:
$
\textbf{\textit{Step1}} \to \textbf{\textit{Step4}} \to \textbf{\textit{Step2}} \to \textbf{\textit{Step3}} \to \textbf{\textit{Step1}} \to \dots
$

\paragraph{Heuristic-Based Controller} In this controller, we create a heuristic Q-function $\hat{Q}_i = \hat{Q}(P_i)$ for each primitive and the controller executes the primitive with the largest $\hat{Q}$. We define 
\begin{equation}
\hat{Q}_i = J_{QP, i} + k_q \sum_{j=1}^4||p_{d, j} - p_j||_2,
\end{equation}
where $J_{QP, i}$ is the QP cost in Equation~\ref{eq:QP} for primitive $P_i$. This represents a trade-off between the pose control and the swing feet control. When more feet are in \textit{Stance} state, $J_{QP, i}$ will be smaller because of the contact constraints. When more feet are in \textit{Swing} state, the foot position error will be smaller since only the swing feet can move to the target position. The heuristic-based controller is able to adapt contact behaviors in a few scenarios but is less robust overall.

\paragraph{End-to-end Learned Controller} We can also directly learn a control policy end-to-end instead of using the hierarchical framework proposed here. However, a purely learned policy on similar scale quadrupeds like the ANYmal and Laikago requires a training sample count on the order of $10^8$ to $10^9$, with careful reward shaping \cite{2020-icra-constraintQuadruped, RoboImitationPeng20, 2019-science-sim2realAnyaml}, while with the hierarchical framework, we use simple reward specification and get good performance at around $10^5$ samples. We train a controller similar to \cite{RoboImitationPeng20}, where the robot is tracking a reference trotting motion. The resulting controller is not as robust as the manually designed trotting controller and fails in most of our testing scenarios. This is consistent with \cite{RoboImitationPeng20}, where it is also found that a purely learned controller is not robust, and a sophisticated adaptation scheme is needed to deal with environmental changes. Given these drawbacks, we do not compare with the end-to-end learned controller.

% \subsection{Hand-designed vs. RL gaits}

% \textbf{Comparing controllers}
% \begin{itemize}
%     \item Zhaoming's RL trotting gaits
%     \item Zhaoming's RL pacing gaits
%     \item Trotting gaits
%     \item Placing gaits
% \end{itemize}

% Measurement: capability and stability

% We will show the controllers following command speed. The current RL control is only trained for x direction walking. The hand-designed one can easily do side walking and turning. We will quantify the stability by initial CoM speed perturbation. We expect the hand-deisgned gait is as good as the learned one, and they could be better. 

% They are also robust for perturbation. They also can directly transfer to experiment without any tuning. Thus, it is not necessary to use RL to learn these primitives or the basic gaits. 

% Explanation: The control architecture builds the basic robustness for the system. It reduces the training time and release the stability concern in learning. 

\subsection{Training Scenarios}
We use a split-belt treadmill to train the policy so that the policy learns to choose different primitives to adapt to changing dynamics. During a new rollout, the speed of the treadmill is sampled from $[-0.3, 0.3]~m/s$, see Figure~\ref{fig:test_scenario} (a). We also randomly pause one side of the treadmill and command the robot to face different orientations. this is shown in Figure~\ref{fig:test_scenario} (b) and (c), where the plywood represents the side of the treadmill that is not moving. This provides a rich set of changing dynamics that the policy must learn to adapt to. Note the policy have no knowledge of the underlying treadmill parameters. The low-level controller is commanded to stay at the origin with a target velocity of $0~m/s$. 
% Some scenarios are depicted in Figure~\ref{fig:test_scenario}. 

% During a new rollout, the orientation of the treadmill is sampled from $[0^{\circ}, 360^{\circ})$ w.r.t the robot yaw orientation.

\begin{figure}
    \centering
    \includegraphics[width=0.95\columnwidth]{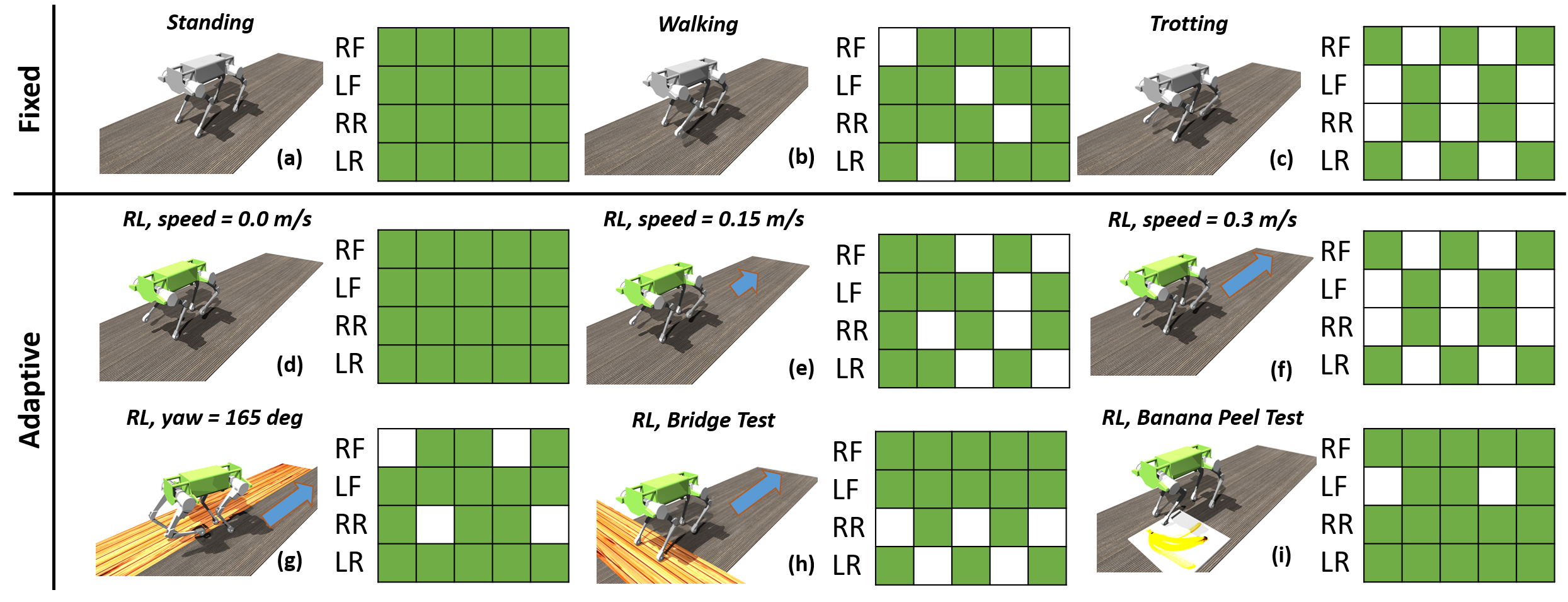}
    \caption{Contact sequence of different high-level controllers under different scenarios. A filled green block indicates that the corresponding foot is in contact with the ground. The three baseline controllers (standing, walking, and trotting) each use a fixed contact sequence for all scenarios, while the learned controller adapts the contact sequence to the scenario.}
    \label{fig:contact_sequence}
    \vspace{-20pt}
\end{figure}

\subsection{Comparison}
\paragraph{Energy} We compare the energy use across different high-level controllers, shown in Figure~\ref{fig:energy_comparison}. The energy is computed as an average of sum square motor torques over ten seconds. 

First, we compare the energy consumption in a scenario shown in Figure~\ref{fig:test_scenario} (a) where the treadmill is moving parallel to the robot in speed range $[0, 0.3]~m/s$. The standing gait is the most energy efficient gait which uses $76.7\%$, $83.3\%$, and $85.7\%$ less energy than the walking, trotting and pacing gaits respectively. The downside is that it can be only used at zero speed. The heuristic and learned (rl) controllers can start with the same lowest energy level and gradually increase the energy level as speed increases. As the treadmill speed reaches $0.2~m /s$, the heuristic controller quickly fails while the learned controller's energy usage is comparable to the walking controller. The walking controller fails when the treadmill speed exceeds $0.2~m/s$ while the learned controller adjusts the primitive such that the energy is the same as the trotting controller. The trotting and pacing controllers are able to cover the full speed range but consume more energy due to unnecessary leg movements.

We then compare the energy consumption in scenarios similar to Figure~\ref{fig:test_scenario} (b) and (c) where only one side of the treadmill is moving at $0.3~m/s$ and the robot is commanded to face different directions. Most of the baseline controllers fail except trotting and pacing. The learned controller consumes on average $40.1\%$ and $50.4\%$ less energy than the trotting and pacing controllers respectively.

\paragraph{Contact Sequence} The energy efficiency of the learned controller is mostly due to adaptive contact planning. Figure~\ref{fig:contact_sequence} shows the contact pattern of baseline and learned controllers. (a), (b), and (c) show the fixed pattern of standing, walking and trotting. Note that they use the same contact sequence for all scenarios. Contrarily, the learned controller adapts contact sequence in different scenarios. When speed increases, the learned controller transients from standing gait to trotting gait, shown in (d), (e), and (f). We highlight the contact pattern in (e), where the learned controller uses a combination of \textbf{\textit{Stand}}, \textbf{\textit{Trot}} and \textbf{\textit{Step}} primitives at speed $0.15~m/s$. In (g), only one belt is moving and we replace the other non-moving belt with plywood for clarity, the learned controller only moves the two right feet, thus more energy efficient compared to trotting or pacing. (h) and (i) are two scenarios not seen in training and the learned controller demonstrates novel contact sequences. 
% Details are explained in the next subsection. 

% \begin{itemize}
%     \item Trotting gait
%     \item Heuristic policy
%     \item Learned policy
% \end{itemize}

% Measurement: Energy level in different speed. If the controller fail on that speed, then no energy is shown. The speed is the treadmill speed for both belts. 

% Figure 1. Energy level vs. speed. for all controllers
% Figure 2. steady state gait selection vs. speed for learned policy

% We will show that at low speed, both the heuristic and learned policy has lower energy than the trotting gait. When speed increased, the heuristic policy failed. The learned controller with change the primitive. When high speed, the learned policy switch between two trotting primitives, which becomes the trotting gait. 

\subsection{Zero-Shot Adaptation to Novel Scenarios}

We test the learned controller in novel scenarios that are not present during training to show that the policy can generalize. 
% Some scenarios are shown in Figure~\ref{fig:test_scenario} (d) and (e). 
A purely learned controller usually overfits the training dynamics and requires the collection of additional data in the target environment to make the adaptation \cite{RoboImitationPeng20}. With our hierarchical framework, the learned controller is able to adapt to these scenarios directly.

% \begin{itemize}
%     \item Trotting gait
%     \item Learned policy
% \end{itemize}

% Test: Differentiate treadmill. Only move one belt at fixed speed 0.3 m/s. Send different command yaw angle. 

% Measurement: Check the energy level and primitive selection. 

% We will show both policy can handle this task, meaning they are robust. Specially the trotting gait, it does not know this task when design the gait. However, the learned gait has more natural motion. The natural could mean what we expect and also mean the lower energy level. 

% Figure 1. Energy level vs. yaw command. for both controller
% Figure 2. steady state gait selection vs. yaw command for learned policy.

% Note: remember to mention MIT Cheetah's work. They can only do longitudinal motion.

\paragraph{Bridge Test}
We test the learned policy on a scenario where the treadmill is placed parallel to the robot while the front legs of the robot are placed on a fixed bridge, see Figure~\ref{fig:test_scenario} (d). This scenario is not present during training while the learned policy is able to adapt and choose not to move the front legs while adjusting the rear legs based on the movement of the treadmill.

\begin{figure}[t!]
\centering
  \begin{minipage}[c]{0.5\textwidth}
  \resizebox{0.98\linewidth}{!}{
  \centering
    \begin{subfigure}[b]{\textwidth}
        \centering
        \includegraphics[width=\textwidth]{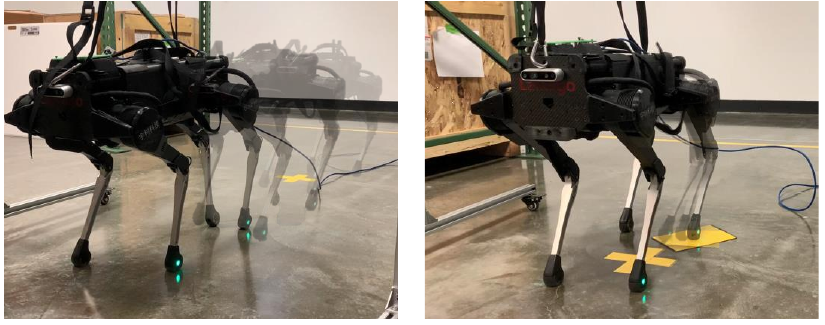}
    \end{subfigure}% \\
    }
\end{minipage}    
  \begin{minipage}[c]{0.4\textwidth}
    \caption{\textbf{Real robot tests}. \\
    \textbf{Left}: Timelapse of forward walking. \\
    \textbf{Right}: The left rear leg is perturbed.
    % with the yellow paper.
    }
 \label{fig:real_robot}
 \end{minipage}
\end{figure}

\paragraph{Banana peel stability test}
We test the robot in another scenario where the treadmill is not moving while a frictionless mat is placed under a foot, represented by a banana peel in Figure~\ref{fig:test_scenario} (e). The only baseline controller that can recover is the heuristic controller, where the slipping foot is adjusted. The learned controller performs similarly to the heuristic controller even though it never sees this situation during training. One can also pass this test with a freeze-joint standing controller, but we emphasize that the high-level contact adjustment can improve the robustness without changing the low-level controller. 

% Test: command is zero, robot stays on the origin. Put a friction-less board under the feet. Four locations: front left foot, rear left foot, between front feet, between rear feet.

% Measurement: The success rate for four tests.

% % Figure 1. Board location respect to the robot. Show four configurations.

% We will show the learned policy has higher success rate, passed 3/4. Where the fixed policies fail all.

% Comment: Surely, one can design a fixed joint standing control that easily passes the test. The point is, with respect to the hand-design fixed policies, the learned policy can increase the robustness in some scenarios. For example, if travel through rough terrain and one leg slips, the learned policy might be able to recover from. We would like to work on these cases in the future. 

% \begin{table}[h!]
% \centering
%  \begin{tabular}{||c c c c||} 
%  \hline
%  Col1 & Col2 & Col2 & Col3 \\ [0.5ex] 
%  \hline\hline
%  1 & 6 & 87837 & 787 \\ 
%  2 & 7 & 78 & 5415 \\
%  3 & 545 & 778 & 7507 \\
%  4 & 545 & 18744 & 7560 \\
%  5 & 88 & 788 & 6344 \\ [1ex] 
%  \hline
%  \end{tabular}
%  \caption{table of comparison stats}
%  \label{tab:comparison}
% \end{table}

\subsection{Sim-to-Real Test}

% \begin{wrapfigure}[24]{r}{0.28\textwidth}
%     \centering
%     \setlength{\fboxsep}{1pt}
%     \begin{subfigure}{.32\textwidth}
%     \includegraphics[width=0.90\columnwidth]{figure/real_robot.pdf}
%     \end{subfigure}
%     \caption{\textbf{Real robot tests}. \\
%     \textbf{Up} Timelapse of the robot walking forward. \\
%     \textbf{Bottom} The leg of the robot is perturbed with the yellow paper.
%     }
%     % \vspace{25pt}
%     \label{fig:real_robot}
% \end{wrapfigure}

We validate the learned controller on the physical robot; snapshots of the experiment are shown in Figure~\ref{fig:real_robot}. Due to the robustness of the low-level controller, we observe that the sim-to-real gap is small compared to other approaches \cite{RoboImitationPeng20}, and the controller is able to perform well without tuning.
%\footnote{The performance is best seen in the supplementary video.}

\paragraph{Walking Forward} To emulate the treadmill, we send the low-level controller command speeds so that the body will move forward and the high-level controller will need to choose primitives to stay balanced. At low speed, the high-level controller first adopts the \textbf{\textit{Stand}} primitive with the body leaning forward; as the robot is close to falling over, other primitives are used to move the leg forward to regain balance. At high speed, the robot mostly uses the \textbf{\textit{Trot}} primitives.

\paragraph{Leg Perturbation} We perturb the legs of the robot by manually pulling them in different directions. The learned controller is able to adopt the corresponding \textbf{\textit{Step}} primitive to move the perturbed leg back to the nominal position while keeping the unperturbed legs still.

% test: in previous test, we show the sim-to-real transfer for hand-design primitive switching. This test is to show the learned policy can be directly replace the hand-designed policy. We expect the policy parameters and low-level control does not need to be changed. 

% Measurement: standing, walking forward, banana peel, turning. All passed. Treadmill is optional.

\section{Conclusion}
\label{sec:conclusion}
We have presented a hierarchical framework that combines model-based control and reinforcement learning. By leveraging the advantages of both paradigms, we obtain a contact-adaptive controller that is more robust and energy efficient than those employing a fixed contact sequence. The learned controller generates novel contact sequences that are generally not produced by either approach alone, at least in the context of real-time control.
We demonstrate our framework using a Laikago quadruped in various challenging scenarios such as walking on a split-belt treadmill with only one side moving or stepping onto a "banana peel." We also validate the controller on the physical robot, finding that sim-to-real transfer is relatively straightforward. We believe this is a promising step toward combining the best features of model-based control and reinforcement learning.

%===============================================================================

% The maximum paper length is 8 pages excluding references and acknowledgements, and 10 pages including references and acknowledgements

\clearpage
% The acknowledgments are automatically included only in the final version of the paper.
% \acknowledgments{TODO: If a paper is accepted, the final camera-ready version will (and probably should) include acknowledgments. All acknowledgments go at the end of the paper, including thanks to reviewers who gave useful comments, to colleagues who contributed to the ideas, and to funding agencies and corporate sponsors that provided financial support.}
\acknowledgments{We thank the NVIDIA Isaac Gym team, especially Viktor Makoviychuk, Lukasz Wawrzyniak, Gavriel State, and many others, for all the kind help they provided in GPU-based simulation and RL training.}

%===============================================================================

% \newpage
\bibliography{example}  % .bib

\begin{thebibliography}{32}
\providecommand{\natexlab}[1]{#1}
\providecommand{\url}[1]{\texttt{#1}}
\expandafter\ifx\csname urlstyle\endcsname\relax
  \providecommand{\doi}[1]{doi: #1}\else
  \providecommand{\doi}{doi: \begingroup \urlstyle{rm}\Url}\fi

\bibitem[Hoyt and Taylor(1981)]{1981-nature-horseGait}
D.~F. Hoyt and C.~R. Taylor.
\newblock Gait and the energetics of locomotion in horses.
\newblock \emph{Nature}, 292\penalty0 (5820):\penalty0 239--240, 1981.

\bibitem[Di~Carlo et~al.(2018)Di~Carlo, Wensing, Katz, Bledt, and
  Kim]{2018-iros-cheetahMPC}
J.~Di~Carlo, P.~M. Wensing, B.~Katz, G.~Bledt, and S.~Kim.
\newblock Dynamic locomotion in the mit cheetah 3 through convex
  model-predictive control.
\newblock In \emph{2018 IEEE/RSJ International Conference on Intelligent Robots
  and Systems (IROS)}, pages 1--9. IEEE, 2018.

\bibitem[Peng et~al.(2020)Peng, Coumans, Zhang, Lee, Tan, and
  Levine]{RoboImitationPeng20}
X.~B. Peng, E.~Coumans, T.~Zhang, T.-W. Lee, J.~Tan, and S.~Levine.
\newblock Learning agile robotic locomotion skills by imitating animals, 2020.

\bibitem[Farshidian et~al.(2017)Farshidian, Jelavic, Satapathy, Giftthaler, and
  Buchli]{2017-humanoids-hyqMPC}
F.~Farshidian, E.~Jelavic, A.~Satapathy, M.~Giftthaler, and J.~Buchli.
\newblock Real-time motion planning of legged robots: A model predictive
  control approach.
\newblock In \emph{2017 IEEE-RAS 17th International Conference on Humanoid
  Robotics (Humanoids)}, pages 577--584. IEEE, 2017.

\bibitem[Winkler et~al.(2018)Winkler, Bellicoso, Hutter, and
  Buchli]{2019-RAL-towr}
A.~W. Winkler, D.~C. Bellicoso, M.~Hutter, and J.~Buchli.
\newblock Gait and trajectory optimization for legged systems through
  phase-based end-effector parameterization.
\newblock \emph{IEEE Robotics and Automation Letters (RA-L)}, 3:\penalty0
  1560--1567, July 2018.
\newblock \doi{10.1109/LRA.2018.2798285}.

\bibitem[Mastalli et~al.(2019)Mastalli, Budhiraja, Merkt, Saurel, Hammoud,
  Naveau, Carpentier, Vijayakumar, and Mansard]{2020-icra-crocoddyl}
C.~Mastalli, R.~Budhiraja, W.~Merkt, G.~Saurel, B.~Hammoud, M.~Naveau,
  J.~Carpentier, S.~Vijayakumar, and N.~Mansard.
\newblock Crocoddyl: An efficient and versatile framework for multi-contact
  optimal control.
\newblock \emph{arXiv preprint arXiv:1909.04947}, 2019.

\bibitem[Tan et~al.(2018)Tan, Zhang, Coumans, Iscen, Bai, Hafner, Bohez, and
  Vanhoucke]{2018-rss-sim2realquadruped}
J.~Tan, T.~Zhang, E.~Coumans, A.~Iscen, Y.~Bai, D.~Hafner, S.~Bohez, and
  V.~Vanhoucke.
\newblock Sim-to-real: Learning agile locomotion for quadruped robots.
\newblock \emph{arXiv preprint arXiv:1804.10332}, 2018.

\bibitem[Hwangbo et~al.(2019)Hwangbo, Lee, Dosovitskiy, Bellicoso, Tsounis,
  Koltun, and Hutter]{2019-science-sim2realAnyaml}
J.~Hwangbo, J.~Lee, A.~Dosovitskiy, D.~Bellicoso, V.~Tsounis, V.~Koltun, and
  M.~Hutter.
\newblock Learning agile and dynamic motor skills for legged robots.
\newblock \emph{Science Robotics}, 4\penalty0 (26), 2019.

\bibitem[Posa et~al.(2014)Posa, Cantu, and Tedrake]{2014-ijrr-direct_traj}
M.~Posa, C.~Cantu, and R.~Tedrake.
\newblock A direct method for trajectory optimization of rigid bodies through
  contact.
\newblock \emph{The International Journal of Robotics Research}, 33\penalty0
  (1):\penalty0 69--81, 2014.

\bibitem[Manchester and Kuindersma(2020)]{2020-springer-zacContactImplicit}
Z.~Manchester and S.~Kuindersma.
\newblock Variational contact-implicit trajectory optimization.
\newblock In \emph{Robotics Research}, pages 985--1000. Springer, 2020.

\bibitem[Mordatch et~al.(2012)Mordatch, Todorov, and
  Popovi{\'c}]{2012-ACM-Igor_contact_invariant_opti}
I.~Mordatch, E.~Todorov, and Z.~Popovi{\'c}.
\newblock Discovery of complex behaviors through contact-invariant
  optimization.
\newblock \emph{ACM Transactions on Graphics (TOG)}, 31\penalty0 (4):\penalty0
  1--8, 2012.

\bibitem[lai()]{laikago}
Laikago website.
\newblock URL
  \url{http://www.unitree.cc/e/action/ShowInfo.php?classid=6&id=1#}.

\bibitem[Boussema et~al.(2019)Boussema, Powell, Bledt, Ijspeert, Wensing, and
  Kim]{2019-RAL-impulseSet}
C.~Boussema, M.~J. Powell, G.~Bledt, A.~J. Ijspeert, P.~M. Wensing, and S.~Kim.
\newblock Online gait transitions and disturbance recovery for legged robots
  via the feasible impulse set.
\newblock \emph{IEEE Robotics and Automation Letters}, 4\penalty0 (2):\penalty0
  1611--1618, 2019.

\bibitem[Xie et~al.(2019)Xie, Clary, Dao, Morais, Hurst, and van~de
  Panne]{2019-CORL-cassie}
Z.~Xie, P.~Clary, J.~Dao, P.~Morais, J.~Hurst, and M.~van~de Panne.
\newblock Learning locomotion skills for cassie: Iterative design and
  sim-to-real.
\newblock In \emph{Proc. Conference on Robot Learning (CORL 2019)}, 2019.

\bibitem[Yu et~al.(2020)Yu, Tan, Bai, Coumans, and Ha]{2020-icra-meta_adapt}
W.~Yu, J.~Tan, Y.~Bai, E.~Coumans, and S.~Ha.
\newblock Learning fast adaptation with meta strategy optimization.
\newblock \emph{IEEE Robotics and Automation Letters}, 5\penalty0 (2):\penalty0
  2950--2957, 2020.

\bibitem[Cully et~al.(2015)Cully, Clune, Tarapore, and
  Mouret]{2015-nature-adapt}
A.~Cully, J.~Clune, D.~Tarapore, and J.-B. Mouret.
\newblock Robots that can adapt like animals.
\newblock \emph{Nature}, 521\penalty0 (7553):\penalty0 503--507, 2015.

\bibitem[Precup(2001)]{2001-precup_thesis-temporal_option}
D.~Precup.
\newblock Temporal abstraction in reinforcement learning.
\newblock 2001.

\bibitem[Li et~al.(2019)Li, Srinivasan, Meng, Yuan, and
  Bohg]{2020-icra-hierachicalManipulation}
T.~Li, K.~Srinivasan, M.~Q.-H. Meng, W.~Yuan, and J.~Bohg.
\newblock Learning hierarchical control for robust in-hand manipulation.
\newblock \emph{arXiv preprint arXiv:1910.10985}, 2019.

\bibitem[Su et~al.(2016)Su, Kroemer, Loeb, Sukhatme, and
  Schaal]{2016-Springer-switch_primitive}
Z.~Su, O.~Kroemer, G.~E. Loeb, G.~S. Sukhatme, and S.~Schaal.
\newblock Learning to switch between sensorimotor primitives using multimodal
  haptic signals.
\newblock In \emph{International Conference on Simulation of Adaptive
  Behavior}, pages 170--182. Springer, 2016.

\bibitem[Peng et~al.(2016)Peng, Berseth, and van~de Panne]{2016-TOG-terrainRL}
X.~B. Peng, G.~Berseth, and M.~van~de Panne.
\newblock Terrain-adaptive locomotion skills using deep reinforcement learning.
\newblock \emph{ACM Trans. Graph.}, 35\penalty0 (4):\penalty0 81:1--81:12, July
  2016.
\newblock ISSN 0730-0301.
\newblock \doi{10.1145/2897824.2925881}.
\newblock URL \url{http://doi.acm.org/10.1145/2897824.2925881}.

\bibitem[Nachum et~al.(2019)Nachum, Ahn, Ponte, Gu, and
  Kumar]{2019-corl-hierachicalsim2real}
O.~Nachum, M.~Ahn, H.~Ponte, S.~Gu, and V.~Kumar.
\newblock Multi-agent manipulation via locomotion using hierarchical sim2real.
\newblock \emph{arXiv preprint arXiv:1908.05224}, 2019.

\bibitem[Peng et~al.(2017)Peng, Berseth, Yin, and van~de
  Panne]{2017-TOG-deepLoco}
X.~B. Peng, G.~Berseth, K.~Yin, and M.~van~de Panne.
\newblock Deeploco: Dynamic locomotion skills using hierarchical deep
  reinforcement learning.
\newblock \emph{ACM Transactions on Graphics (Proc. SIGGRAPH 2017)},
  36\penalty0 (4), 2017.

\bibitem[Nachum et~al.(2018)Nachum, Gu, Lee, and
  Levine]{2018-nips-hierachicalRL}
O.~Nachum, S.~S. Gu, H.~Lee, and S.~Levine.
\newblock Data-efficient hierarchical reinforcement learning.
\newblock In \emph{Advances in Neural Information Processing Systems}, pages
  3303--3313, 2018.

\bibitem[Jain et~al.(2019)Jain, Iscen, and
  Caluwaerts]{2019-arxiv-hierarchicalQuadruped}
D.~Jain, A.~Iscen, and K.~Caluwaerts.
\newblock Hierarchical reinforcement learning for quadruped locomotion.
\newblock \emph{arXiv preprint arXiv:1905.08926}, 2019.

\bibitem[Tsounis et~al.(2020)Tsounis, Alge, Lee, Farshidian, and
  Hutter]{2020-icra-deepgait}
V.~Tsounis, M.~Alge, J.~Lee, F.~Farshidian, and M.~Hutter.
\newblock Deepgait: Planning and control of quadrupedal gaits using deep
  reinforcement learning.
\newblock \emph{IEEE Robotics and Automation Letters}, 5\penalty0 (2):\penalty0
  3699--3706, 2020.

\bibitem[Mnih et~al.(2015)Mnih, Kavukcuoglu, Silver, Rusu, Veness, Bellemare,
  Graves, Riedmiller, Fidjeland, Ostrovski, et~al.]{2015-nature-dqn}
V.~Mnih, K.~Kavukcuoglu, D.~Silver, A.~A. Rusu, J.~Veness, M.~G. Bellemare,
  A.~Graves, M.~Riedmiller, A.~K. Fidjeland, G.~Ostrovski, et~al.
\newblock Human-level control through deep reinforcement learning.
\newblock \emph{nature}, 518\penalty0 (7540):\penalty0 529--533, 2015.

\bibitem[Van~Hasselt et~al.(2016)Van~Hasselt, Guez, and
  Silver]{2016-aaai-doubleDQN}
H.~Van~Hasselt, A.~Guez, and D.~Silver.
\newblock Deep reinforcement learning with double q-learning.
\newblock In \emph{Thirtieth AAAI conference on artificial intelligence}, 2016.

\bibitem[Fujimoto et~al.(2018)Fujimoto, Van~Hoof, and Meger]{2018-icml-td3}
S.~Fujimoto, H.~Van~Hoof, and D.~Meger.
\newblock Addressing function approximation error in actor-critic methods.
\newblock \emph{arXiv preprint arXiv:1802.09477}, 2018.

\bibitem[Gangapurwala et~al.(2020)Gangapurwala, Mitchell, and
  Havoutis]{2020-icra-constraintQuadruped}
S.~Gangapurwala, A.~Mitchell, and I.~Havoutis.
\newblock Guided constrained policy optimization for dynamic quadrupedal robot
  locomotion.
\newblock \emph{IEEE Robotics and Automation Letters}, 5\penalty0 (2):\penalty0
  3642--3649, 2020.

\bibitem[NVIDIA(2020)]{2020-IsaacGym}
NVIDIA.
\newblock \emph{Isaac Gym - Preview Release}, 2020.
\newblock URL \url{https://developer.nvidia.com/isaac-gym}.

\bibitem[Liang et~al.(2020)Liang, Handa, Van~Wyk, Makoviychuk, Kroemer, and
  Fox]{2020-arxiv-phyx1}
J.~Liang, A.~Handa, K.~Van~Wyk, V.~Makoviychuk, O.~Kroemer, and D.~Fox.
\newblock In-hand object pose tracking via contact feedback and gpu-accelerated
  robotic simulation.
\newblock \emph{arXiv preprint arXiv:2002.12160}, 2020.

\bibitem[Chebotar et~al.(2019)Chebotar, Handa, Makoviychuk, Macklin, Issac,
  Ratliff, and Fox]{2019-icra-phyx2}
Y.~Chebotar, A.~Handa, V.~Makoviychuk, M.~Macklin, J.~Issac, N.~Ratliff, and
  D.~Fox.
\newblock Closing the sim-to-real loop: Adapting simulation randomization with
  real world experience.
\newblock In \emph{2019 International Conference on Robotics and Automation
  (ICRA)}, pages 8973--8979. IEEE, 2019.

\end{thebibliography}

% dont need this in submission time yet
\newpage
\appendix

\section{Linearized Centroidal Dynamics}
\label{sec:dynamics derivation}

The dynamics is similar to \cite{2018-iros-cheetahMPC} with a few modifications. The general centroidal dynamics is
\begin{equation}
 \ddot{p}_{\textit{body}}=\frac{\sum_{i=1}^{4} f_{i}}{m}-g,
 \end{equation}	
\begin{equation}
    \mathbf{I} \dot{\omega}=\sum_{i=1}^{4} p_{i} \times f_{i},
\end{equation}
where $\ddot{p}_{\textit{body}}$ is the base linear acceleration, $f_i$ is the ground reaction force on each foot, and $m$, $g$ are the mass and gravity vector respectively. The $\bf{I}$ is the mass inertia, $\bf{\dot{\omega}}$ is the derivative of angular velocity. The $p_i$ is the foot position respect to the base. All variables are represented in the world frame. We ignore the Coriolis force $\omega \times (\bf{I} \omega)$ since it does not contribute significantly to the dynamics of the robot.

We use the small angular assumption to linearize the dynamics. The robot's orientation is expressed as a vector of Z-Y-X Euler angles $\Theta=[\phi \; \theta \; \psi]^{\top}$, where $\phi$ is roll, $\theta$ is pitch, and $\psi$ is yaw. For small values of roll and pitch $(\phi, \theta)$, the angular velocity is approximated by

\begin{equation}
\omega\approx \mathbf{R}_{z}(\psi) \dot{\Theta},
\end{equation}
where
$$
\mathbf{R}_{z}(\psi) = \left[\begin{array}{ccc}
\cos (\psi) & -\sin (\psi) & 0 \\
\sin (\psi) & \cos (\psi) & 0 \\
0 & 0 & 1
\end{array}\right]
$$
is the rotation matrix of yaw. The inertia matrix in the world frame can be approximated by
\begin{equation}
\mathbf{I}\approx \mathbf{R}_{z}(\psi)\;_{\mathcal{B}} \mathbf{I} \; \mathbf{R}_{z}^\top(\psi),
\end{equation}
where $_{\mathcal{B}} \mathbf{I}$ is the inertia matrix in body frame.

The linearized dynamic is 
\begin{equation}
\ddot{q} = \mathbf{M} f-\begin{bmatrix} g\\ 0_3
\end{bmatrix},
\end{equation}
where
\begin{equation}
\mathbf{M}=\left[\begin{array}{ccc}
1_{3} / m & \ldots & 1_{3} / m \\
\mathbf{R}_{z}^\top\; _{\mathcal{B}}\mathbf{I}^{-1}\left[p_{1}\right]_{\times} & \ldots & \mathbf{R}_{z}^\top\; _{\mathcal{B}}\mathbf{I}^{-1}\left[p_{4}\right]_{\times}
\end{array}\right].
\end{equation}

\section{Q-Learning Algorithm}
We use DQN like algorithm to train our high-level policy. Details are shown in Algorithm~\ref{alg:DQN}.
\label{sec:DQN}

\begin{algorithm}[H]
\SetAlgoLined
 initialization Q-function parameters $\theta_1. \theta_2$ for $Q_{\theta_1}, Q_{\theta_2}$, empty replay buffer $D$ \;
 set target network parameters $\theta_{targ, 1}, \theta_{targ,2} \leftarrow \theta_1, \theta_2$ for $Q_{\theta_{targ, 1}}, Q_{\theta_{targ, 2}}$ \;
 \While{not done}{
  observe current state $s$ \;
  sample action $a$ based on Q-function\;
  observe next state $s'$, reward $r$ and done signal $d$\;
  store $(s, a, r, d, s')$ in replay buffer $D$\;
  \If{d is True or time limit reached}{
   reset environment\;
   }
  \If{time to update}{
    \For{$j = 1, 2, \dots \text{number of update}$}{
        sample batch of transition data $B=\{s, a, r, d, s'\}$\;
        compute $a' = \argmax_a Q_\theta(s', a)$\;
        compute target $q_{targ} = r + (1-d)\gamma \min_{i = 1,2}( Q_{\theta_{targ, i}}(s', a'))$\;
        update $\theta_1, \theta_2$ by taking gradient descent w.r.t the objective function $\frac{1}{|B|}\sum_{(s,a,r,d,s') \in B}((Q_{\theta_1}(s,a)-q_{targ})^2 + (Q_{\theta_2}(s,a)-q_{targ})^2)$ \;
        \If{$j \mod 2 = 1$}{
            $\theta_{targ, 1} \leftarrow \rho\theta_{targ, 1}+(1-\rho)\theta_1$ \;
            $\theta_{targ, 2} \leftarrow \rho\theta_{targ, 2}+(1-\rho)\theta_2$ \;
        }
    }
  }
 }
 \caption{Q Learning}
 \label{alg:DQN}
\end{algorithm}

\end{document}